\documentclass[sigconf]{acmart}

\copyrightyear{2027}
\acmYear{2027}
\setcopyright{cc}
\setcctype{by}
\acmConference[ASPDAC '27]{32nd Asia and South Pacific Design Automation Conference}{January 25--28, 2027}{Tokyo, Japan}
\acmBooktitle{32nd Asia and South Pacific Design Automation Conference (ASPDAC '27), January 25--28, 2027, Tokyo, Japan}
\acmDOI{10.1145/3840404.3848406}
\acmISBN{979-8-4007-2923-2/2027/01}
\usepackage{booktabs}
\usepackage{array}
\usepackage{multirow}

\newcommand{\pp}{\,pp}
\newcommand{\method}{T3R}

\begin{document}

\title{T3R: Training-Free Two-Stage Token Refinement Towards Efficient and Robust VLA Models}

\author{Mohammed Hassan}
\affiliation{%
  \institution{The University of Texas at Austin}
  \city{Austin}
  \state{Texas}
  \country{United States}
}
\email{mohassan@utexas.edu}

\author{Zhenyang Chen}
\affiliation{%
  \institution{Georgia Institute of Technology}
  \city{Atlanta}
  \state{Georgia}
  \country{United States}
}
\email{chenzy086@gmail.com}

\author{Zheng Wang}
\affiliation{%
  \institution{University of Illinois Urbana-Champaign}
  \city{Urbana}
  \state{Illinois}
  \country{United States}
}
\email{zhengw10@illinois.edu}

\author{Zhixin Zhu}
\affiliation{%
  \institution{Hong Kong University of Science and Technology}
  \city{Hong Kong}
  \country{Hong Kong}
}
\email{cindy.zhu@live.cn}

\author{Tianlong Chen}
\affiliation{%
  \institution{University of North Carolina at Chapel Hill}
  \city{Chapel Hill}
  \state{North Carolina}
  \country{United States}
}
\email{tianlong@cs.unc.edu}

\author{Yingyan (Celine) Lin}
\affiliation{%
  \institution{Georgia Institute of Technology}
  \city{Atlanta}
  \state{Georgia}
  \country{United States}
}
\email{celine.lin@gatech.edu}

\author{Chaojian Li}
\affiliation{%
  \institution{Hong Kong University of Science and Technology}
  \city{Hong Kong}
  \country{Hong Kong}
}
\email{chaojian@ust.hk}
\renewcommand{\shortauthors}{Hassan et al.}

\begin{abstract}
\begingroup
\setlength{\emergencystretch}{2em}
Vision-Language-Action (VLA) models enable end-to-end robot control, but inference is increasingly expensive: each decision processes hundreds of visual tokens, often across multiple cameras, leading to substantial computation while allowing irrelevant background content to distract action prediction under distribution shift. We argue that these issues stem from two complementary forms of redundancy: many visual tokens are spatially irrelevant, while attention over the remaining tokens is often misallocated. Based on this observation, we propose T3R, a Training-Free Two-Stage Token Refinement framework. In the first stage, segmentation-guided token pruning removes irrelevant visual patches to reduce computation and eliminate distracting context. In the second stage, saliency-guided attention biasing reallocates attention toward action-critical visual regions and language instructions within the retained tokens. Together, pruning produces a cleaner candidate set, while attention biasing further emphasizes the most decision-relevant evidence, yielding a better accuracy-efficiency frontier than either stage alone. On LIBERO-Spatial, T3R removes 85\% of scene-camera visual tokens while essentially matching the unpruned success rate, and improves success by up to 7 percentage points under out-of-distribution object-placement perturbations. The timing comparison attributes 7.8\,ms per inference call to pruning and attention-bias application. As input resolution increases, LLM-forward profiling shows up to a $3.3\times$ speedup over the unpruned baseline. Evaluations across LIBERO, SIMPLER, and RoboTwin with three VLA backbones show the strongest accuracy-efficiency benefits in multi-camera settings. These results demonstrate that training-free token refinement is a practical approach to efficient and robust VLA inference.
\par
\endgroup
\end{abstract}



\maketitle

\section{Introduction}
Vision-Language-Action (VLA) models~\cite{rt2,pi0,openvla} provide a unified
interface from pixels and language to robot actions, but this generality carries
a substantial inference burden. At every decision step, the model propagates a
long visual sequence through a large language backbone. In a typical
OpenVLA-style setup, a $224\times224$ image produces $256$ visual
patches~\cite{siglip}, and multi-camera inputs multiply this token load before
any action token is predicted. Attending over such long sequences at every
control step is expensive and increases latency, which is detrimental to
real-time control and can further induce observation--action distribution shift
through stale observations and delayed actuation~\cite{sail}.

\begin{figure*}[!t]
\centering
\includegraphics[width=0.86\textwidth]{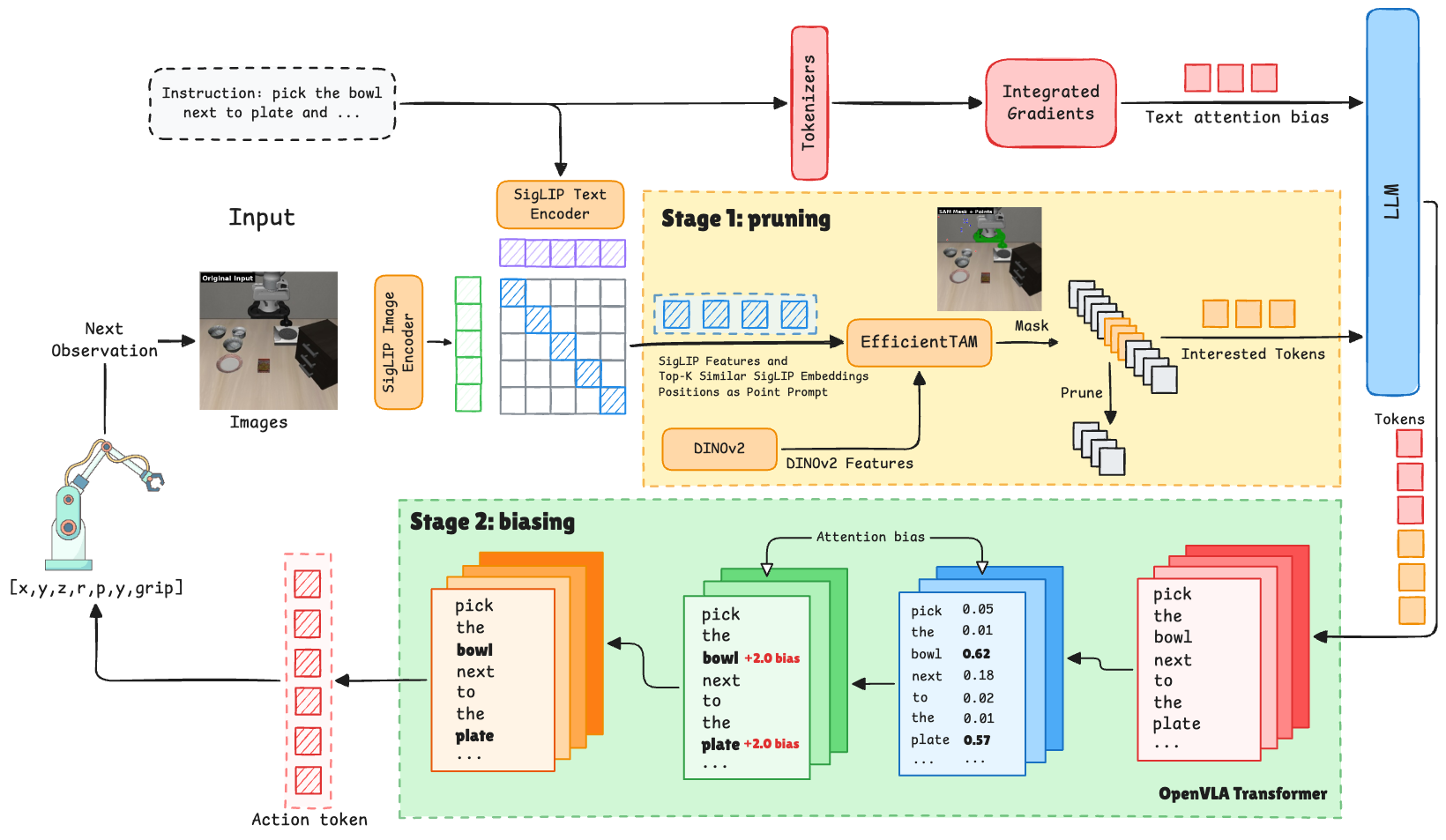}
\Description{A closed-loop robot-control diagram. Camera observations and the
instruction feed image and text encoders. In the first stage, similarity-selected
image patches and DINOv2 features guide EfficientTAM to generate a mask; only
selected visual tokens enter the language model. In the second stage, Integrated
Gradients assigns importance to instruction words, illustrated by bowl and plate,
and attention biases emphasize those words and selected visual tokens. The model
outputs action tokens, the robot executes the actions, and a new observation
starts the next iteration.}
\caption{Two-stage \method{} pipeline. \textbf{Stage~1 (pruning):} POS-weighted
SigLIP text--patch similarity localizes the target; the top-scoring patch
positions (with SigLIP/DINOv2 features) prompt EfficientTAM to produce an object
mask, which is converted to a patch-level pruning decision while preserving
original RoPE positions, so only the ``interested tokens'' enter the LLM.
\textbf{Stage~2 (biasing):} Integrated Gradients over the instruction identifies
salient words; an additive attention bias ($\beta{=}2.0$) is injected into the
middle decoder layers for action-token queries, steering attention toward salient
text tokens and retained visual patches before the action chunk is decoded.}
\label{fig:pipeline}
\end{figure*}

Many visual tokens are weakly coupled to the language
instruction~\cite{fastv,prumerge}. These tokens can diffuse attention and distract
action prediction, especially under viewpoint or object-placement shifts.
Effective refinement must both remove irrelevant tokens and reinforce
instruction-relevant evidence among those retained.

Redundancy acts at two coupled stages: \emph{token-set construction} and
\emph{attention allocation}. Pruning can discard critical cues, while attention
steering retains the full token cost and noisy input. We therefore prune first
and refine attention only over the retained sequence.

Building on this view, we introduce \method{}, a training-free two-stage token
refinement framework. Stage~1 performs segmentation-guided token pruning with
position preservation, so compression happens \emph{before} the VLA decoder and
every transformer layer benefits from a shorter input. Stage~2 compensates for
pruning-induced information loss by injecting task relevance into the retained
sequence: saliency-guided attention biasing in intermediate layers steers
action-token queries toward instruction-salient words and retained task-relevant
regions.

We evaluate \method{} on OpenVLA-OFT~\cite{openvlaoft}, CogACT~\cite{cogact},
and $\pi_{0.5}$~\cite{pi05} across LIBERO~\cite{libero}, SIMPLER~\cite{simpler},
and RoboTwin~\cite{robotwin}. Across these systems, multi-camera settings show
greater pruning headroom than the single-camera zero-shot setting. This trend
suggests that overlapping visual evidence creates opportunities for token reduction.

\noindent\textbf{Contributions.}
\begin{itemize}
\item We propose \method{}, a training-free two-stage framework that couples
segmentation-guided, position-preserving token pruning with saliency-guided
attention biasing, jointly optimizing token selection and attention allocation
in a single inference pass.
\item On LIBERO-Spatial with OpenVLA-OFT, \method{} removes 85\% of scene-camera
tokens while retaining 96\% success, reduces VLA inference latency by 14\%,
and improves OOD success by up to 7\pp{}. Resolution-scaling profiling shows
up to a $3.3\times$ LLM-forward speedup over the unpruned baseline.
\item We evaluate \method{} across VLA backbones and camera setups. The observed
accuracy-efficiency trends are consistent with greater sensor redundancy
providing more headroom for token pruning.
\end{itemize}

\section{Related Work}
\noindent\textbf{Vision-Language-Action models.}
VLA methods cast robot control as conditional sequence modeling over
observations, instructions, and actions. RT-1/RT-2~\cite{rt1,rt2} established
large-scale vision-language pretraining for manipulation, while OpenVLA and
OpenVLA-OFT~\cite{openvla,openvlaoft} improved openness and adaptation
efficiency. Parallel efforts including $\pi_0$/$\pi_{0.5}$~\cite{pi0,pi05},
CogACT~\cite{cogact}, and GR00T~\cite{groot} scale model capacity and control
fidelity. Inference still relies on propagating long visual token sequences
through heavy backbones, leaving efficiency and robustness under perturbation as
open bottlenecks.

\noindent\textbf{Token-efficient inference.}
Token reduction has been studied in vision transformers and multimodal LLMs via
pruning, merging, and routing~\cite{tome,dynamicvit,fastv,prumerge}. In embodied
settings, EfficientVLA, LightVLA, and TeamVLA~\cite{efficientvla,lightvla,teamvla}
adapt these ideas with task-aware scoring and runtime compression; FastV~\cite{fastv}
prunes visual tokens after an early decoder layer; ADP~\cite{adp} performs
action-aware dynamic pruning; and VLA-Cache~\cite{vlacache} reuses KV-cache
entries across frames. \method{} targets the harder \emph{decoder-entry} regime:
tokens are removed before \emph{any} decoder layer, benefiting all layers, but
selection is irreversible and lacks internal attention guidance. \method{}
addresses this by coupling external vision-language grounding with post-pruning
attention repair. \method{} and VLA-Cache are complementary (spatial vs.\
temporal redundancy).

\noindent\textbf{Segmentation priors and attribution steering.}
Promptable segmentation models such as SAM/SAM2 and efficient
variants~\cite{sam,sam2,efficienttam} provide a strong prior for locating
instruction-relevant regions; we use segmentation as an \emph{intermediate}
computational prior rather than a perception output. Attribution methods,
especially Integrated Gradients~\cite{ig}, estimate token contribution without
parameter updates; since raw attention is not always a faithful importance
indicator~\cite{attn_not_expl,patro}, we use external relevance signals to bias
middle-layer action-token attention, following the spirit of inference-time
intervention~\cite{atlas}.

\noindent\textbf{Positioning.}
To our knowledge, \method{} is the first training-free approach to prune 85\%
of scene-camera tokens \emph{before} decoder entry while retaining success
within 1\pp{} of the unpruned baseline on LIBERO-Spatial.
External grounding and attention repair benefit every decoder layer without
retraining.

\section{Method}
\method{} follows a two-stage pipeline (Fig.~\ref{fig:pipeline}):
segmentation-guided token pruning (Sec.~\ref{sec:prune}) removes spatially
redundant tokens, then saliency-guided attention biasing (Sec.~\ref{sec:bias})
sharpens attention on the most important tokens that remain.

\subsection{Segmentation-Guided Token Pruning}
\label{sec:prune}
We remove visually irrelevant patches before they reach the VLA backbone,
reducing effective sequence length while preserving tokens that carry
task-relevant evidence. We first estimate patch-level relevance with the SigLIP
vision-language encoder, which yields a text--patch similarity map, then refine
it into an object mask with a SAM-based segmenter.

\noindent\textbf{Task-guided patch scoring.}
Given an instruction $\ell$ and an image $I\in\mathbb{R}^{224\times224\times3}$,
SigLIP partitions the image into $N{=}256$ patches with features
$V\in\mathbb{R}^{N\times d}$ and tokenizes the instruction into
$T\in\mathbb{R}^{M\times d}$. Natural-language instructions mix nouns, spatial
relations, and function words that are not equally informative for grounding.
We therefore compute a part-of-speech (POS)-weighted patch-relevance score
\begin{equation}
s_i=\frac{\sum_{j=1}^{M} w_j\,\cos(v_i,t_j)}{\sum_{j=1}^{M} w_j},
\end{equation}
where $t_j$ is the embedding of token $j$, $w_j$ its POS-dependent weight, and
$\cos(\cdot,\cdot)$ cosine similarity. This softly emphasizes object nouns and
descriptive adjectives while down-weighting verbs and prepositions, without
supervision or architectural changes.

\noindent\textbf{Segmentation-based mask generation.}
Relevance scores localize the target coarsely but lack precise boundaries. We
refine them with EfficientTAM-S~\cite{efficienttam}, which generates masks from
sparse point prompts. Because nearly uniform similarity scores make top-$K$
prompt selection unstable, we adopt an \emph{oversample-cluster-refine}
strategy: select the $K_{\text{init}}{=}50$ highest-scoring patches, apply DBSCAN
($\epsilon{=}2.0$ in patch units) on their 2D grid coordinates, retain the
largest spatial cluster, and take its $K{=}8$ top-scoring patches as positive
prompts plus $K_{\text{neg}}{=}3$ lowest-scoring patches as negative prompts.
This spatial coherence reduces mask flicker across frames. A patch is retained
if the mask covers more than 30\% of its pixels ($\tau{=}0.3$). SigLIP text
embeddings are cached per instruction; masks are updated asynchronously every
five VLA calls.

\begin{figure}[!htb]
\centering
\includegraphics[width=0.80\columnwidth]{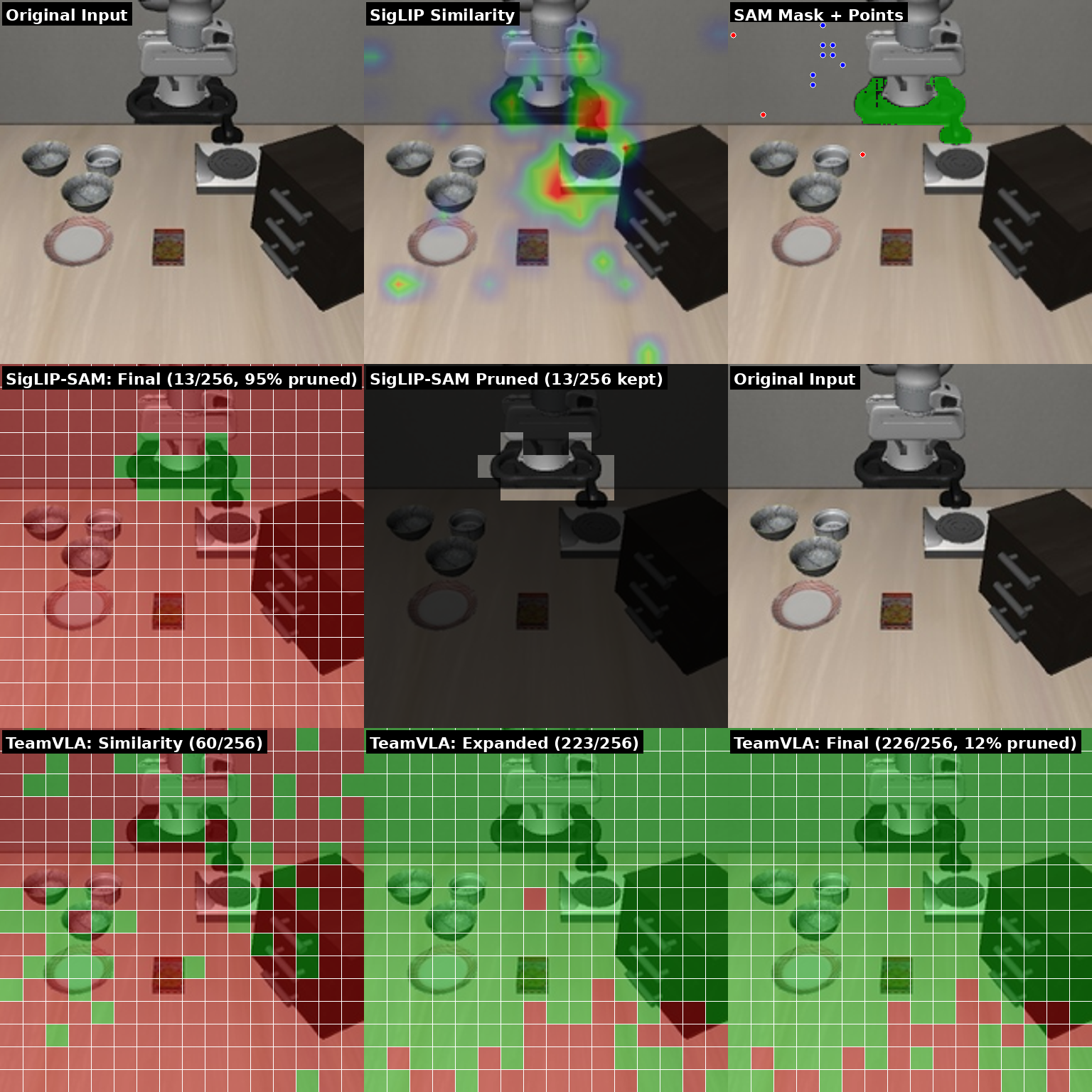}
\Description{A three-by-three panel comparison of patch selection in a tabletop
scene containing a robot gripper, bowls, a plate, and a ramekin. The top row shows
the original image, a colored similarity heatmap, and a green segmentation mask
with positive and negative point prompts. The middle row shows a small retained
region around the target, the image with discarded regions darkened, and the
original image for comparison. The bottom row shows TeamVLA's similarity,
expansion, and final merging masks, which retain much more background. Green
patches are retained and red patches are discarded.}
\caption{Qualitative pruning on a sparse LIBERO-Spatial scene (``pick up the bowl
between the plate and the ramekin''). \textbf{Top:} input, POS-weighted SigLIP
similarity, and the EfficientTAM mask with its point prompts. \textbf{Middle:}
\method{} keeps only 13/256 patches (95\% pruned), tightly localized on the
target. \textbf{Bottom:} TeamVLA's similarity+expansion+merge retains 226/256
(12\% pruned), leaving most of the background in the sequence through the early
decoder layers.}
\label{fig:prune}
\end{figure}

\noindent\textbf{Position-preserving pruning.}
The input layout is
\[
[\textsc{bos},\,v_{1:256},\,v'_{1:256},\,q_{\text{proprio}},\,t_{1:L}]
\]
with sequential RoPE positions~\cite{rope}. Naively re-indexing after pruning shifts
downstream tokens to positions never seen in pretraining and collapses success
to near zero in our tests. We instead assign each retained patch its
\emph{original} position ID: for the retained index set $S$ the sequence becomes
\[
[0,\{i{+}1\}_{i\in S},257\text{--}512,513,514,\dots],
\]
preserving the pretrained positional structure and visual--language relationships.

\noindent\textbf{Scheduling.}
Segmentation runs \emph{asynchronously}, with updates scheduled every five VLA
inference calls. Action inference reuses the latest completed mask without
waiting for an update. On suites where segmentation is ambiguous, an optional
similarity-union rule additionally retains patches above the 75th percentile
of the similarity scores, trading a little compression for recall.

\subsection{Saliency-Guided Attention Biasing}
\label{sec:bias}
Pruning inevitably removes scene information. To mitigate the loss of
task-relevant signal, we steer attention toward salient instruction tokens and
retained task-relevant patches within the reduced sequence.

\noindent\textbf{Attribution via Integrated Gradients.}
We adopt Integrated Gradients (IG)~\cite{ig}, which attributes a prediction to
input tokens by integrating gradients along a straight-line path from a neutral
baseline to the actual input and satisfies the completeness axiom. At the start
of each task we compute token-level IG on the embedding layer, interpolating
between a pad-token baseline and the input in $n_{\text{steps}}{=}5$ steps; the
normalized attribution $\alpha\in\mathbb{R}^{L}$ designates the top 30\% of
tokens as salient. The attributions are cached and reused across all evaluation
episodes of that task.

\noindent\textbf{Attention-bias injection.}
We add the attribution as an additive bias to the attention logits of the middle
decoder layers:
\begin{equation}
a_{ij}=\frac{q_i\!\cdot\!k_j}{\sqrt{d}}
+\beta\, w_j\,\mathbf{1}[j\in S_{\text{bias}}]\,\mathbf{1}[i\in Q_{\text{action}}],
\end{equation}
where $\beta$ is a scalar strength, $w_j$ the bias weight at key position $j$,
$S_{\text{bias}}$ the union of salient text positions and retained visual patch
positions, and $Q_{\text{action}}$ the action-prediction queries. The bias is
applied only to action-token queries and restricted to middle layers (8--23),
preserving early feature extraction and late output formatting. Biased positions
are drawn from (i) salient text tokens weighted by their IG scores and (ii)
retained visual patches weighted by SigLIP similarity scaled by $0.5$. This forms
a two-level funnel: pruning removes clearly irrelevant tokens; biasing further
differentiates among those that remain.

\begin{figure*}[t]
\centering
\includegraphics[width=0.77\textwidth]{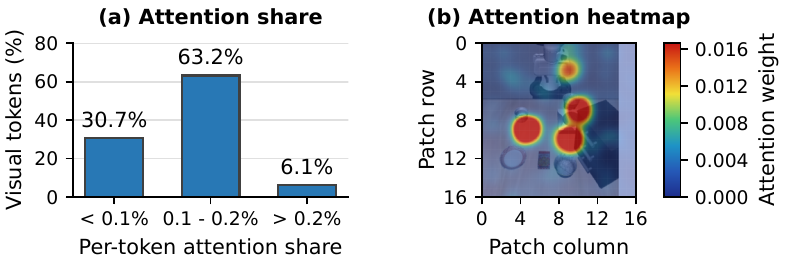}
\Description{Two panels summarize visual-token attention. The left bar chart
shows that 30.7 percent of tokens receive less than 0.1 percent attention each,
63.2 percent receive 0.1 to 0.2 percent, and 6.1 percent receive more than
0.2 percent. The right panel overlays a patch-grid heatmap on the scene. Most
patches have low attention, with a few high-attention regions near the gripper
and target bowls. The color scale runs from blue for low attention to red for
high attention.}
\caption{Most visual tokens are near-inert. (a)~Across all 32 decoder layers,
93.9\% of visual tokens receive $<$0.2\% of the action queries' attention each.
(b)~Per-patch action-query attention concentrates on a few task-relevant patches
(here the gripper and target bowls), while the background receives negligible
weight.}
\label{fig:attention}
\end{figure*}

\section{Experiments}
We test three hypotheses: (H1)~segmentation-guided pruning substantially reduces
scene-camera tokens while maintaining success; (H2)~fewer tokens lower inference
cost; and (H3)~the full T3R pipeline improves robustness under out-of-distribution
object-placement shifts. We then examine cross-backbone generalization and
sensor redundancy.

\noindent\textbf{Setup.}
Our primary study uses OpenVLA-OFT~\cite{openvlaoft} (SigLIP ViT-SO400M +
LLaMA-2-7B, LoRA rank 32, chunk size 8, dual camera $\Rightarrow$ 512 visual
tokens/call) on LIBERO~\cite{libero}, a MuJoCo tabletop-manipulation suite of
four task families (Spatial, Object, Goal, Long). Spatial uses a 220-step horizon.
Saliency uses IG with $n_{\text{steps}}{=}5$, top-$k$ ratio $0.3$, $\beta{=}2.0$,
layers 8--23, and a $0.5$ visual scaling. Unless noted we evaluate 10
episodes/task (seed 7 nominal, seed 42 OOD) on a single NVIDIA L40S (48\,GB).
For generalization we additionally port \method{} to CogACT~\cite{cogact} on
SIMPLER~\cite{simpler} (single third-person camera) and to
$\pi_{0.5}$~\cite{pi05} on RoboTwin~\cite{robotwin} (three cameras), covering a
$1\times$/$2\times$/$3\times$ spread in camera count. Baselines are the unpruned
model, TeamVLA~\cite{teamvla}, ADP~\cite{adp}, and FastV~\cite{fastv}; each
baseline runs at its native operating point, and we additionally force ADP/FastV
into \method{}'s high-reduction regime where noted. Reported reductions refer to
scene-camera tokens removed before decoder entry.

\subsection{\method{} Prunes Aggressively While Maintaining Success (H1)}
\begin{table}[!htb]
\centering
\caption{LIBERO-Spatial (OpenVLA-OFT). ``Reduction'' is scene-camera tokens
pruned before decoder execution; latency is per VLA inference call.
$^{*}$ADP native operating point
(67\% of steps $\times$ 50\%/step $\approx$ 33.5\% reduction).}
\label{tab:libero_main}
\footnotesize
\setlength{\tabcolsep}{5pt}
\begin{tabular}{@{}lccc@{}}
\toprule
Method & Reduction & SR (\%) & Latency (ms)\\
\midrule
Baseline (no pruning) & 0\% & 97 & 107.5\\
TeamVLA~\cite{teamvla} & $\sim$15\% & 94 & 102.6\\
FastV~\cite{fastv} ($k{=}1$, $r{=}0.85$) & 85\% & 14 & --\\
ADP~\cite{adp} & 85\% & 66 & --\\
ADP~\cite{adp}$^{*}$ & 33.5\% & 97 & --\\
\textbf{\method{} (ours)} & \textbf{85\%} & \textbf{96} & \textbf{92.5}\\
\bottomrule
\end{tabular}
\end{table}

Table~\ref{tab:libero_main} shows that \method{} achieves approximately
$5.7\times$ the scene-camera token-reduction rate of TeamVLA (85\% vs.\
$\sim$15\%), while achieving a higher success rate (96\% vs.\ 94\%).
At the same 85\% reduction, ADP and FastV fall to 66\% and 14\%; native ADP
preserves accuracy at only 33.5\% reduction. The retained budget adapts to task
difficulty: \method{} keeps 9--15 patches in sparse scenes and up to 121 in
clutter, versus TeamVLA's fixed $\sim$203--233, avoiding a worst-case token count.

Across all four LIBERO suites (10 tasks $\times$ 10 episodes each), \method{}
matches or beats TeamVLA on latency everywhere and on success on 3/4 suites,
matching the unpruned baseline within 1\pp{} on Spatial (96 vs.\ 97) and Goal
(98 vs.\ 99) at roughly 88--93\,ms/call vs.\ the baseline's 105--108\,ms. The Object
suite is the exception (82\% vs.\ TeamVLA's 86\%), reflecting segmentation
failures discussed in the failure analysis.

\subsection{Most Visual Tokens Are Negligible}
\label{sec:negligible}
For each decoder layer we sum attention mass from every action-token query to
every visual key and express each token's share as a percentage of the total,
averaging over 32 layers, 10 LIBERO-Spatial tasks, and 3 timesteps/task.
\textbf{93.9\% of visual tokens receive less than 0.2\% of action-query attention
each}. In late layers (24--31), 50--75\% of tokens fall below 0.1\%. This
shows most tokens are near-inert, but safe pre-decoder removal still depends on
selecting the retained set well.

\subsection{\method{} Reduces Computational Cost (H2)}
Profiling on an L40S across 11 token counts (8--256 patches) reveals an
approximately linear trend, $0.040\,\text{ms/token}+53.3\,\text{ms}$
($r{=}0.97$). At $224$\,px, 85\% scene-camera token reduction lowers VLA
inference latency from 107.5 to 92.5\,ms per call: a 14\% reduction or
$1.16\times$ speedup. Each call produces an eight-action chunk, not one
environment step.

Comparing LLM-only savings ($58.9-36.1$\,ms) with reported VLA-call savings
($107.5-92.5$\,ms) yields a 7.8\,ms residual per call, attributed to pruning and
attention-bias application and already included in 92.5\,ms. This accounting
assumes comparable timing conditions; it does not isolate individual components.
Segmentation (54\,ms per update, every five VLA calls) runs asynchronously.
IG is computed once per task and reused across its episodes; a 1\,s cost
corresponds to approximately 0.45\,ms per environment step over ten full
220-step episodes.

\noindent\textbf{Resolution scaling and hardware implications.}
The effective per-token marginal cost rises from $0.040$\,ms at sequence length
$\sim$200 to $0.064$\,ms at $\sim$500, marking the transition from a
bandwidth-bound to a compute-bound attention regime. Because \method{} fixes the
pruning \emph{ratio} rather than an absolute token count, its retained sequence
stays in the cheaper bandwidth-bound regime as resolution grows, while the
unpruned baseline enters the quadratic compute-bound regime.
LLM-forward profiling shows up to a $3.3\times$ speedup at 768\,px.
For accelerator co-design the implication is concrete: pre-decoder pruning shifts
the operating point back toward the memory-bound regime that current LLM
inference hardware handles most efficiently. Using the measured LLM cost on an
L40S and adding a fixed 48.6\,ms non-LLM cost, we project end-to-end inference
latencies of $100/115/132$\,ms at $384/512/768$\,px for \method{},
versus the baseline's $144/182/327$\,ms and ADP's $109/145/238$\,ms, a
$2.47\times$ projected speedup over baseline at 768\,px, versus ADP's
$1.37\times$. These projections do not independently measure refinement
overhead at each resolution.

\begin{figure}[!b]
\centering
\includegraphics[width=\columnwidth]{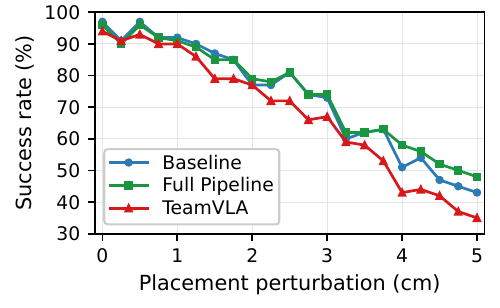}
\Description{A line chart plots success rate against placement perturbation
from zero to five centimeters. Blue circles represent the unpruned baseline,
green squares represent the full T3R pipeline, and red triangles represent
TeamVLA. The curves begin close together at high success rates and generally
decline as perturbation increases. Beyond about two centimeters, the green
curve is above or equal to the blue curve, while the red curve declines most
strongly.}
\caption{OOD robustness under object-placement perturbation (21 noise levels,
10 episodes/task). \method{} (Full Pipeline) tracks the unpruned baseline for
small shifts and pulls ahead beyond ${\sim}2$\,cm, leading by up to 7\pp{} at
4\,cm, while TeamVLA degrades fastest. The crossover is consistent with pruning
reducing misleading background context.}
\label{fig:ood}
\end{figure}

\begin{table*}[t]
\centering
\caption{Cross-backbone generalization. \textbf{Left:} CogACT on SIMPLER
(single camera), average SR (\%); each keep ratio corresponds to the respective
method's native operating point. \textbf{Right:} $\pi_{0.5}$ on RoboTwin
(three cameras), SR (\%) vs.\
per-camera token budget (base $=33.3$). Best per column/row in bold.}
\label{tab:gen}
\footnotesize
\renewcommand{\arraystretch}{1.15}
\begin{minipage}[t]{0.46\textwidth}
\centering
\setlength{\tabcolsep}{6pt}
\begin{tabular}{@{}lcc@{}}
\toprule
Method (keep ratio) & Google Robot & WidowX/Bridge\\
\midrule
Baseline (100\%) & 73.2 & 10.4\\
ADP~\cite{adp} (75\% / $\sim$87\%) & 71.0 & \textbf{16.7}\\
FastV~\cite{fastv} ($\sim$60\%) & \textbf{73.3} & 15.6\\
TeamVLA~\cite{teamvla} (31.25\%) & 69.5 & 12.5\\
\textbf{\method{} (ours, 60\%)} & 67.7 & 10.4\\
\bottomrule
\end{tabular}
\end{minipage}\hfill
\begin{minipage}[t]{0.50\textwidth}
\centering
\setlength{\tabcolsep}{7pt}
\begin{tabular}{@{}cccccc@{}}
\toprule
tok/cam & keep & \method{} & ADP & FastV & TeamVLA\\
\midrule
64  & 25\%   & \textbf{13.3} & \textbf{13.3} & 11.7 & 3.3\\
96  & 37.5\% & \textbf{25.0} & 13.3 & 23.3 & 11.7\\
128 & 50\%   & \textbf{33.3} & 21.7 & 18.3 & 15.0\\
160 & 62.5\% & \textbf{38.3} & 25.0 & 28.3 & 15.0\\
\bottomrule
\end{tabular}
\end{minipage}
\end{table*}

\subsection{\method{} Improves OOD Robustness (H3)}
We apply Gaussian positional noise to object placements (0--5\,cm; 21 levels,
10 ep/task). \method{} tracks the baseline within 1--2\pp{} for small
perturbations, matches or exceeds it at ${\geq}2$\,cm, and leads by up to
7\pp{} at 4\,cm ($+4.8$\pp{} average at ${\geq}4$\,cm); TeamVLA degrades
fastest (35\% at 5\,cm vs.\ our 48\%). These full-pipeline gains are consistent
with pruning reducing misleading background context and attention refinement
emphasizing task-relevant evidence.

\subsection{Generalization and Sensor Redundancy}
\label{sec:generalize}
To test whether \method{}'s benefit transfers across backbones and sensor suites,
we port it to CogACT on SIMPLER (single third-person camera) and to $\pi_{0.5}$
on RoboTwin (three cameras). Both evaluations are zero-shot with no fine-tuned
checkpoint, so absolute success is low; we focus on \emph{relative} behavior
versus the unpruned base and competing pruners at matched budgets.

\noindent\textbf{Single-camera (CogACT / SIMPLER).}
Table~\ref{tab:gen} (left) reports CogACT on the SIMPLER Google-Robot (11 tasks,
189 episodes per method) and WidowX/Bridge (4 tasks, 96 episodes per method)
suites, each using one third-person view. Here \method{} with a 60\% keep ratio matches
base on Bridge and trails by 5.5\pp{} on Google-Robot, while methods retaining
more tokens (ADP, FastV) fare better. With one view, little cross-sensor
redundancy remains, so aggressive pruning removes tokens the policy still needs.

\noindent\textbf{Multi-camera ($\pi_{0.5}$ / RoboTwin).}
For $\pi_{0.5}$ on RoboTwin (three cameras), we evaluate each method/budget on
five tasks with 12 episodes each (60 total; Table~\ref{tab:gen}, right).
Success rates are computed from pooled counts and rounded once.
\method{} achieves the highest or joint-highest success among
the evaluated pruners across the reported budgets. At a 50\% keep ratio it
matches baseline (both 33.3\%); at 62.5\% it reaches 38.3\%,
exceeding baseline by 5.0\pp{} and the best competing pruner by 10\pp{}.
ADP, FastV, and TeamVLA remain below baseline at these two budgets.

\noindent\textbf{Takeaway.}
Across the evaluated backbones and benchmarks, \method{} shows greater pruning
headroom in multi-camera settings. This cross-system trend suggests that
overlapping visual evidence provides opportunities for token reduction.

\subsection{Ablations}
We isolate each component on LIBERO-Spatial (10 tasks $\times$ 10 episodes).
Prune-only reaches 90\% at 85\% scene-camera token reduction, bias-only 98\%, and full
\method{} 96\%, versus a 97\% baseline. Biasing recovers 6\pp{} of pruning loss;
bias-only exceeding baseline indicates diffuse unpruned attention. In the
bias-only layer sweep (no pruning), middle-layer biasing (8--23) reaches 98\%,
versus 95\% for early, 92\% for late, and 93\% for
all layers, consistent with early feature distortion and late output disruption.

\noindent\textbf{Why \method{} outperforms prior pruners.}
\method{} prunes before all 32 decoder layers ($\sim$37 scene tokens retained),
while TeamVLA carries 512 tokens through layers 0--15, allowing irrelevant
context into representations before merging. Segmentation preserves coherent
object regions and RoPE geometry rather than disjoint similarity samples;
IG-guided attention repair then offsets pruning loss.

\noindent\textbf{Hyperparameter robustness.}
One hyperparameter set served all 10 LIBERO-Spatial tasks and all four suites,
without per-task tuning. Sweeping bias strength, active layers, saliency top-$k$,
and pruning threshold changed SR by 1--6\pp{}; DBSCAN $\epsilon$ and SAM point
counts follow the fixed patch-grid geometry.

\subsection{Design Study: When and How to Steer Attention}
A pilot on ECoT-OpenVLA-7B~\cite{ecot} (30 offline LIBERO-Spatial demos, L2 action
error) crossed text-token IG with image-token Grad-CAM~\cite{gradcam}, reasoning
with action phases, and positive with negative biasing. Action-phase biasing cuts
error by 5.4\% ($p{=}0.062$), while reasoning-phase biasing raises it by 9.3\%
($p{=}0.018$). Text IG outperforms image Grad-CAM, and negative biasing never
helps. These results support positive text-IG steering during action generation:
the policy under-attends relevant tokens, so biasing can recover pruning loss.

\subsection{Failure Analysis and Limitations}
\method{} underperforms TeamVLA on LIBERO-Object (82\% vs.\ 86\%). These tasks
place a source object into a visually dominant destination (e.g., a basket), so
SigLIP similarity splits between the two and EfficientTAM segments the wrong
region or produces an overly narrow mask. Multi-phase LIBERO-Long tasks (e.g., ``put the
bowl in the drawer and close it'') pose a related issue: the relevant region
shifts mid-episode, but similarity is computed from the full instruction and does
not adapt to the current phase, so pruning may mask regions that become critical
later. More broadly, (i)~Stage~1 depends on promptable segmentation as an
intermediate prior and inherits its failure modes on merged or coarse
boundaries; (ii)~IG adds a one-time per-task cost that may not amortize over short
task runs or meet strict initialization deadlines; and (iii)~our single-camera zero-shot results
(Sec.~\ref{sec:generalize}) show the method offers little benefit when the sensor
suite carries little redundancy. Phase-aware re-prompting and camera-adaptive
budgets are natural extensions.

All evaluations use simulation; physical-robot control and asynchronous execution
with real sensor and actuation delays remain unvalidated. Hardware profiling uses
only an NVIDIA L40S, with no energy measurements or edge-device tests. Latency
improvements alone therefore do not establish energy savings or edge performance.
Occlusion and lighting changes may degrade segmentation and discard task-relevant
patches. We have not isolated these effects on task success or compared
alternative segmenters.

\section{Conclusion}
We presented \method{}, a training-free two-stage framework that couples
segmentation-guided token pruning with saliency-guided
attention biasing. On LIBERO-Spatial with OpenVLA-OFT, it removes 85\% of
scene-camera visual tokens while retaining success within 1\pp{} of baseline,
reduces VLA inference latency, and improves OOD robustness by up to 7\pp{}.
Evaluations on CogACT and $\pi_{0.5}$ across SIMPLER and RoboTwin suggest that
overlapping multi-camera evidence provides more headroom for token pruning.
From a design perspective, the value of \method{} lies in \emph{where} it
acts. By discarding irrelevant tokens before decoder entry, every transformer
layer operates on the shorter sequence, so the retained tokens stay in the
cheaper bandwidth-bound regime even as resolution grows, yielding a
projected speedup of up to
$2.47\times$ at $768$\,px. Mask updates run asynchronously, while IG computation
is amortized across each task's episodes. Because the pipeline is training-free and leaves the
underlying VLA weights unchanged, it applies directly to existing pretrained
policies. Together, these properties make
task-conditioned, pre-decoder sparsification a practical, first-class knob for
co-designing efficient embodied inference systems. Across scenes, its adaptive
budget retains 9--15 patches in sparse cases and up to 121 in clutter, while
middle-layer biasing recovers 6\pp{} of the prune-only accuracy loss. Together
with the 93.9\% attention-sparsity result, these measurements explain why
aggressive pruning remains viable and when additional evidence must be retained.

\enlargethispage{3\baselineskip}
\begin{acks}
This research was supported by the NVIDIA Academic Grant Program using NVIDIA DGX Spark and NVIDIA Jetson AGX Thor.
\end{acks}

\clearpage
\bibliographystyle{ACM-Reference-Format}
\bibliography{references}

\end{document}